\documentclass[10pt,twocolumn,letterpaper]{article}
\usepackage[margin=0.75in,columnsep=0.25in]{geometry}
\usepackage[T1]{fontenc}
\usepackage{newtxtext,newtxmath}
\usepackage[numbers,sort&compress]{natbib}
\usepackage{booktabs}
\usepackage{tabularx}
\usepackage{array}
\usepackage{xcolor}
\usepackage{tikz}
\usetikzlibrary{positioning,arrows.meta,fit,backgrounds}
\usepackage{enumitem}
\usepackage[hidelinks]{hyperref}
\usepackage{url}
\usepackage{graphicx}
\graphicspath{{figs/}}
\usepackage[skip=4pt plus 1pt]{parskip}  
\setlist{nosep,leftmargin=*}
\usepackage[hang,flushmargin]{footmisc}              
\usepackage{titlesec}                                
\titleformat*{\section}{\normalfont\large\bfseries\raggedright\hyphenpenalty=10000\exhyphenpenalty=10000}
\titleformat*{\subsection}{\normalfont\normalsize\bfseries\raggedright\hyphenpenalty=10000\exhyphenpenalty=10000}
\titleformat{\paragraph}[runin]{\normalfont\normalsize\bfseries}{}{0pt}{}
\titlespacing*{\paragraph}{0pt}{2pt plus 1pt}{0.5em}  
\usepackage[font=small,labelfont=bf]{caption}        

\newcommand{\act}[1]{\texttt{#1}}
\newcommand{\agentname}{WebFovea}

\title{\textbf{\agentname: When the Model Is Right but the Click Is Wrong}\\[3pt]
\large Reliable Round Trips for Vision-Based Web Agents on Live Websites}
\author{Jiangang Han\\
\small Independent Researcher\\
\small \texttt{jiangangh@gmail.com}}
\date{}

\begin{document}
\twocolumn[
\begin{@twocolumnfalse}
\maketitle
\begin{abstract}
\noindent We present \agentname, a vision-based web agent that placed 2nd in the WebRetriever Challenge 2026~\cite{dong2026webretrieverlargescalecomprehensivebenchmark,webretriever_challenge} with a final score of 57.0 out of 100. The challenge evaluates agents end to end on Protocol~III of the WebRetriever benchmark~\cite{dong2026webretrieverlargescalecomprehensivebenchmark}: starting from an entry URL on a live website, the agent must operate the site's own interface and return a verifiable answer. A capable multimodal large language model (LLM) is necessary for this, but not sufficient. The model's decisions reach the browser through the \emph{harness}, the code between the model and the page. At every step, four things must go right: the model's reply must be parsed into the intended action, the action must take effect on the page, the result must be reported back accurately, and the model must be shown the information it needs. On real websites, many of the failures we observed occurred at one of these four stages rather than in the model's reasoning. A coordinate-space mismatch placed every click at 3/4 of its intended coordinates; actions on native dropdowns, inside iframes, and in text boxes failed silently; and self-generated chat-template tokens contaminated 4.9\% of task episodes. \agentname{} hardens each stage and surrounds the loop with guardrails that keep the agent within the rules and its budget. The four-stage view does not depend on the model, although some individual fixes do. Because we used the same model in all four submissions, the rise of our official hidden-set score from 31.0 to 57.0 reflects changes to the harness, up to run-to-run variance on live sites. We describe the design, the evidence for each component (including negative results), a failure analysis, the limitations, and a roadmap that includes routing different steps to different models.

\smallskip\noindent\textbf{Code:} \url{https://github.com/jianganghan/WebFovea}
\end{abstract}
\vspace{1em}
\end{@twocolumnfalse}
]

\section{Introduction}

Much of the information people and organizations need lives on public websites: statistical portals, regulatory filings, rankings, and dashboards. Retrieving it usually means navigating menus, setting filters, reading tables and charts, and only then reading off a value. Web agents that automate this are useful only if their answers can be trusted. The WebRetriever benchmark~\cite{dong2026webretrieverlargescalecomprehensivebenchmark} evaluates such agents on 1,550 tasks on 800 real, live websites. Its Protocol~III, the challenge track, asks for end-to-end retrieval: navigate from an entry URL, operate the page, and return a single answer that can be checked exactly. Success requires \emph{both} a correct answer and a trajectory that actually reaches the evidence. The benchmark's own results show why both are needed: navigation success alone is an insufficient predictor of task completion~\cite{dong2026webretrieverlargescalecomprehensivebenchmark}.

This calls for \textbf{grounding at two levels}:
\begin{itemize}
\item \textbf{Answer-level grounding.} The answer must come from the authoritative page, reached through its interface during this session, not from parametric memory, a search engine, or a hand-crafted API call. Our logs show why this matters. After hitting a CAPTCHA, an early version of our agent answered from memory; in another task, the model wrote an entire fictitious multi-step trajectory in a single reply and submitted an answer at step~1 (\S\ref{sec:parsing}). Web data also drifts: by the time we evaluated, some of the 100 public reference answers no longer matched the live sites (\S\ref{sec:failure}). In such cases, only the trajectory and the final on-screen evidence can show whether the agent's answer was right.
\item \textbf{Action-level grounding.} Every action must land on the intended pixel or element, and its effect must be perceived correctly. For a vision-based agent, this is where many of our failures turned out to lie.
\end{itemize}

\begin{figure*}[t]
\centering
\begin{tikzpicture}[
  ln/.style={draw, rounded corners=2pt, fill=blue!7, align=flush center, font=\footnotesize, text width=3.9cm, minimum height=1.25cm, inner sep=3pt},
  ep/.style={draw, rounded corners=3pt, align=flush center, font=\footnotesize, text width=2.45cm, minimum height=3.6cm, inner sep=3pt},
  arr/.style={-{Stealth[length=2mm]}, thick}]
\node[ep] (model) at (0,0) {\textbf{Multimodal LLM}\\(decides the next action)\\[3pt]here: \texttt{claude-opus-4-6}\\[3pt]could be: per-step routing to large or small models, rules, classical CV};
\node[ep] (page) at (15.0,0) {\textbf{Live website}\\[3pt]Chrome over CDP\\[3pt]iframes, native widgets, canvas charts, obfuscated fonts};
\node[ln] (pars) at (5.0,1.05) {\textbf{Parsing}\\keep the first action only; strip self-generated tokens};
\node[ln] (exec) at (10.2,1.05) {\textbf{Execution}\\coordinate-space alignment; element-level \act{select}/\act{fill}; iframe DOM fallback};
\node[ln] (feed) at (10.2,-1.05) {\textbf{Feedback}\\DOM + child-frame fingerprints; read-back; text hash for tooltips};
\node[ln] (obs) at (5.0,-1.05) {\textbf{Observation}\\screenshot size cap; \act{hover}; \act{find\_text}; \act{read\_text} (de-obfuscated)};
\draw[arr] (model.east |- pars) -- node[above, font=\scriptsize]{reply} (pars);
\draw[arr] (pars) -- node[above, font=\scriptsize]{action} (exec);
\draw[arr] (exec) -- node[above, font=\scriptsize]{effect} (page.west |- exec);
\draw[arr] (page.west |- feed) -- node[below, font=\scriptsize]{change} (feed);
\draw[arr] (feed) -- node[below, font=\scriptsize]{report} (obs);
\draw[arr] (obs) -- node[below, font=\scriptsize]{view} (model.east |- obs);
\begin{scope}[on background layer]
\node[draw, dashed, rounded corners=6pt, fill=orange!6, inner sep=9pt, fit=(pars)(exec)(feed)(obs),
      label={[font=\footnotesize, yshift=-1pt]above:\textbf{Guardrails}: closed action set $\cdot$ hard constraints $\cdot$ budgets enforced by the executor $\cdot$ no task retries $\cdot$ model-call retries}] (gov) {};
\end{scope}
\end{tikzpicture}
\caption{Each step is a round trip through four stages. The model's reply is parsed into an action and executed on the page; the page's reaction comes back as feedback and as the next view. Guardrails bound the whole loop.}
\label{fig:links}
\end{figure*}
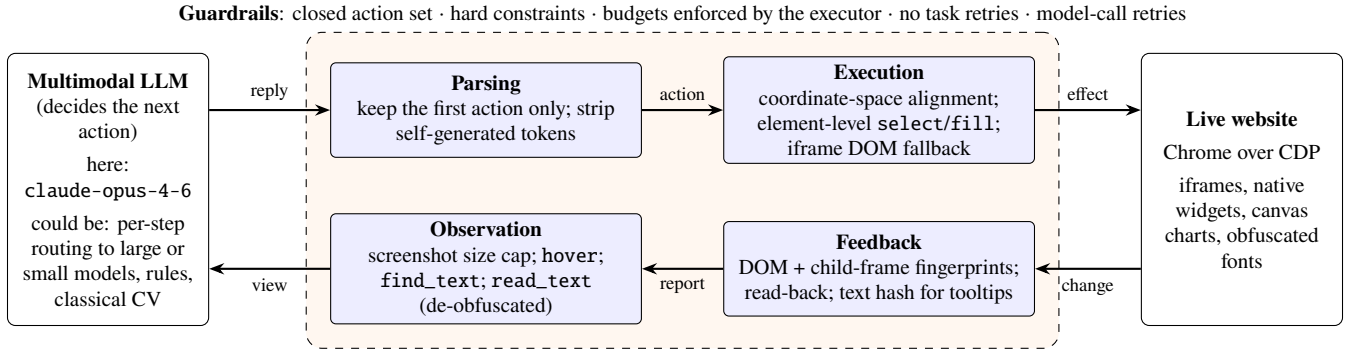

Both levels depend on the \emph{harness}: everything between the model and the browser. We find it useful to view each step as a round trip that passes through four stages (Figure~\ref{fig:links}, \S\ref{sec:links}). The model's reply is turned into an action (\emph{parsing}), the action is carried out on the page (\emph{execution}), the page's reaction is reported back (\emph{feedback}), and the next view shows the model what it needs (\emph{observation}). When one stage breaks, even a strong model fails in ways that look like reasoning errors: it clicks empty space, retypes text that never arrived, or abandons an action that actually worked. The model is right, but the click is wrong.

The challenge restricted teams to a list of approved models, and we used the same one throughout, so the changes in our score reflect changes to the harness rather than to the model. This is not an argument for keeping the model fixed: stronger and more specialized models help, and an agent may route different steps to different models, rules, or classical vision tools (\S\ref{sec:future}). But whichever model decides, its decisions still pass through the same four stages.

\paragraph{Contributions.}
(1)~A simple way to analyze vision-based web agents: \textbf{each step is a round trip through four stages} (parsing, execution, feedback, and observation), \textbf{with guardrails around the loop}. The view doubles as a diagnostic checklist: for any failure, ask at which stage it broke. (2)~\textbf{\agentname}\footnote{Named after the fovea, the small central part of the retina that gives sharp vision. Just as the eye shifts its gaze to bring each detail onto the fovea, the agent examines a page one view at a time until the evidence is clear.}, an agent built on this view, which placed 2nd in the WebRetriever Challenge 2026 with a score of 57.0 out of 100. Across four submissions, harness changes raised its score from 31.0 to 57.0, and it lost no points in the organizers' manual review. (3)~\textbf{Evidence for each stage}: paired evaluations, case studies, and negative results. (4)~\textbf{Design principles}, an explicit account of the limitations, and a prioritized roadmap, including routing steps to different models and adding classical computer-vision tools.

\section{Related Work}

\paragraph{Web agent benchmarks.}
Early benchmarks used self-hosted sites (WebArena~\cite{zhou2024webarena}) or offline snapshots (Mind2Web~\cite{deng2023mind2web}); OSWorld~\cite{xie2024osworld} extends evaluation to full desktop environments. WebVoyager~\cite{he2024webvoyager} moved to live websites with multimodal agents. WebRetriever~\cite{dong2026webretrieverlargescalecomprehensivebenchmark} scales live-site evaluation to 1,550 tasks on 800 websites spanning many industries and languages. It defines three protocols: basic navigation, navigation with operational documentation, and end-to-end retrieval. It also introduces NavEval, which judges navigation from network requests, URL trajectories, actions, and the final screenshot, and agrees with human experts 91.2\% of the time. Its central finding, that navigation success alone is an insufficient predictor of task completion, motivates the challenge's focus on Protocol~III. Table~\ref{tab:leaderboard} lists the Protocol~III baselines on the benchmark leaderboard. They were evaluated on a different task set with human verification, so they are not directly comparable to challenge scores.

\paragraph{Agent paradigms.}
Web agents follow two broad paradigms. \emph{Vision-based} agents take screenshots of the page and act through pixel coordinates. Examples include Claude computer use~\cite{anthropic2024computeruse}, Gemini computer use~\cite{gemini25}, and UI-TARS~\cite{qin2025uitars}, a GUI agent model trained end to end; SeeClick~\cite{cheng2024seeclick} showed that better GUI grounding improves such agents. \emph{DOM-based} agents serialize the page into indexed elements. Examples include Browser Use~\cite{browseruse2024}, Agent-E with DOM distillation and change observation~\cite{abuelsaad2024agente}, and SeeAct, which grounds GPT-4V's action descriptions in HTML candidates~\cite{zheng2024seeact}. Set-of-Mark prompting~\cite{yang2023setofmark} sits between the two by overlaying element labels on the screenshot.

\paragraph{The interface between model and environment.}
Most of this work improves the model or the way the page is represented to it. Closest in spirit to ours is SWE-agent~\cite{yang2024sweagent}, which showed that the design of the agent--computer interface strongly affects how well a language model solves software-engineering tasks. We study the analogous question for vision-based web agents: the path a given model's decisions take to the page and back.

\begin{table}[tb]
\centering\small
\caption{Protocol~III baselines on the WebRetriever leaderboard~\cite{webretriever_leaderboard} (success rate, human-verified). $^\dagger$Model not on the challenge's approved list.}
\label{tab:leaderboard}
\begin{tabular}{@{}llr@{}}
\toprule
Agent & Model & SR (\%) \\
\midrule
Claude Computer Use & claude-4-8-opus$^\dagger$ & 47 \\
Kimi Agent & Kimi K3$^\dagger$ & 28 \\
Google Computer Use & gemini-2.5-pro & 21 \\
Claude Computer Use & claude-4-5-sonnet & 16 \\
Browser Use & gpt-4o & 11 \\
Agent-E & gpt-4o & 9 \\
UI-TARS-1.5 & 7B & 8 \\
SeeAct & gpt-4o & 6 \\
\bottomrule
\end{tabular}
\end{table}

\begin{table*}[t]
\centering\small
\caption{The four stages and the guardrails: failures observed in our runs and the components that address them.}
\label{tab:links}
\begin{tabularx}{\textwidth}{@{}>{\raggedright\arraybackslash}p{2.5cm}>{\raggedright\arraybackslash}X>{\raggedright\arraybackslash}X@{}}
\toprule
Stage & What goes wrong (example from our runs) & Components \\
\midrule
Parsing (\S\ref{sec:parsing}) & chat-template tokens typed into search boxes (4.9\% of episodes); an invented trajectory with a fake final answer & first-action truncation; stripping special tokens and stray trailing markup \\
Execution (\S\ref{sec:execution}) & every click at 3/4 of its intended coordinates; native dropdowns that clicks cannot operate; keystrokes lost to the wrong focus; clicks swallowed inside iframes; an Enter the prompt promised but the executor never sent & coordinate-space alignment; \act{select}; \act{fill}; Enter on newline; iframe DOM fallback \\
Feedback (\S\ref{sec:feedback}) & a successful iframe click reported as ``no detectable change'', steering the model away; failed inputs invisible in the screenshot & DOM and child-frame fingerprints; notes naming the element hit; \act{fill} read-back; tooltip text hash \\
Observation (\S\ref{sec:observation}) & values that appear only on hover; targets far down long pages; digits hidden by custom fonts; verification tools not used when needed & screenshot size cap; \act{hover}; \act{find\_text}; \act{read\_text}; rule-based triggers \\
Guardrails (\S\ref{sec:guardrails}) & shortcuts through URLs and data APIs; endless repetition; tasks lost to a crashed worker or a failed API call & closed action set; hard constraints; executor budgets; scroll breaker; retries and dispatch \\
\bottomrule
\end{tabularx}
\end{table*}

\section{Task Setting}
\label{sec:setting}

\paragraph{Protocol~III.} Each task gives a natural-language instruction and an entry URL. The agent must operate the live site and call \act{finished(content=...)} with its answer. Tasks span document extraction, form interaction, multi-source comparison, complete data retrieval, and chart analysis~\cite{dong2026webretrieverlargescalecomprehensivebenchmark}. There is no partial credit. Final scores combine answer correctness with a manual review of trajectories and code.

\paragraph{Submission and runtime.} Teams submit code to a private repository, and the organizers run it on 100 hidden tasks in a designated sandbox, as the challenge rules require~\cite{webretriever_challenge}. The agent drives a Chrome browser over the Chrome DevTools Protocol (CDP) at a 1920$\times$1080 viewport, within a fixed time budget per evaluation run. Each team gets at most one evaluation per day over a seven-day window, and only an aggregate score is returned. Models are called through an OpenAI-compatible API and must come from an approved list; the newest approved Anthropic model was Claude~4.6. We used \texttt{claude-opus-4-6} throughout.

\paragraph{Validity criteria.} A score counts only if the run meets three criteria~\cite{webretriever_challenge}. (i)~\emph{Step-by-step interaction}: every action follows from the current page; the agent may not bypass the interface, jump to precomputed results, or enumerate data APIs. (ii)~\emph{Real-time autonomy}: no pre-stored or cached answers. (iii)~\emph{Source compliance}: answers come from the designated site, without search engines or third-party data. Tasks may not be retried or reset once started; only individual steps may be retried (for example, after a failed model call).

\paragraph{Starting point.} We built \agentname{} on the code template that the organizers gave every team: a trimmed-down copy of the WebRetriever repository's reference implementation~\cite{dong2026webretrieverlargescalecomprehensivebenchmark}, an agent written for UI-TARS~1.5~\cite{qin2025uitars}. For brevity, we also call this template the \emph{reference implementation}. Appendix~\ref{app:inventory} lists which parts we kept and which are ours. We developed on the 100 publicly released Protocol~III tasks, split into a 70-task development set (\emph{dev70}) and a 30-task holdout set (\emph{holdout30}) (\S\ref{sec:expsetup}).

\section{The \agentname{} Agent}
\label{sec:method}

\subsection{Four stages and guardrails}
\label{sec:links}

At every step, the model looks at what it is shown and decides on one action. It never touches the browser directly: the harness carries its decision to the page and carries the page's reaction back. We view this as a round trip through four stages, and a step succeeds only if all four work (Figure~\ref{fig:links}, Table~\ref{tab:links}):

\begin{itemize}
\item \textbf{Parsing:} what the model writes is what the harness understands.
\item \textbf{Execution:} the action the model intends is the one that happens on the page.
\item \textbf{Feedback:} the model is told what actually happened.
\item \textbf{Observation:} what the model needs to see is in front of it.
\end{itemize}
Around the loop, \textbf{guardrails} keep the agent within the rules and within its budget, and keep infrastructure faults from ending a task. They are not a stage of the round trip, but they decide which round trips are allowed and how many.

\paragraph{How \agentname{} implements the loop.} \agentname{} keeps the observe--think--act loop of the reference implementation. At each step, the model receives the task, a description of the action space and the hard constraints, the full text history of its earlier thoughts and actions, the five most recent screenshots, and, when relevant, a one-sentence \emph{feedback note} about the previous action. It replies with a \texttt{Thought} and one \texttt{Action}. The agent is \emph{vision-based}: the model receives no DOM or accessibility tree and acts through pixel coordinates. DOM access is confined to the executor and to two narrow text channels: the feedback notes and \act{read\_text}. Table~\ref{tab:actions} lists the final action space. A single executor function is the only place where the agent can affect the browser.

\begin{table}[tb]
\centering\small
\caption{Final action space. Budgets are enforced by the executor, not only stated in the prompt.}
\label{tab:actions}
\begin{tabularx}{\columnwidth}{@{}>{\raggedright\arraybackslash}p{0.40\columnwidth}>{\raggedright\arraybackslash}X@{}}
\toprule
Action & Purpose (per-task budget) \\
\midrule
\act{click}, \act{left\_double}, \act{right\_single}, \act{drag}, \act{scroll}, \act{wait} & Basic mouse and wait actions (inherited) \\
\act{hover} & Reveal tooltips / chart values \\
\act{select(box, option)} & Native \texttt{<select>} dropdowns \\
\act{fill(box, content)} & Focus, clear, type, read back \\
\act{type(content)} & Keystrokes; trailing \texttt{\textbackslash n} presses Enter \\
\act{hotkey(key)} & Page-level keys; browser-level keys blocked \\
\act{find\_text(text)} & Scroll text into view; returns no text ($\le$10) \\
\act{read\_text(near)} & Read DOM text to verify a value ($\le$5) \\
\act{back()} / \act{reload()} & History back ($\le$15) / reload ($\le$2) \\
\act{click\_many(points)} & Image-grid CAPTCHA only ($\le$6) \\
\act{finished(content)} & Submit answer (empty if unreachable) \\
\bottomrule
\end{tabularx}
\end{table}

The rest of this section goes through the stages in the order a step passes through them, and ends with the guardrails.

\subsection{Parsing: reading the model's reply as intended}
\label{sec:parsing}

\begin{figure}[t]
\centering
\includegraphics[width=\columnwidth]{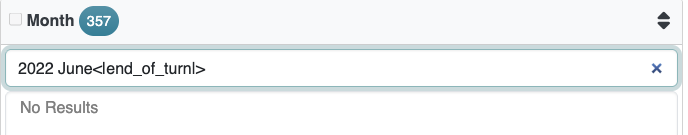}
\caption{A parsing failure: a self-generated chat-template token typed into a data portal's filter box (CSO PxStat; task: CPI for June 2022). The site filters on the literal string and returns no results. The model eventually diagnosed the problem in its own reasoning but kept re-emitting the token and ran out of its 40 steps.}
\label{fig:token}
\end{figure}

We found two output failures that had nothing to do with reasoning. (i)~\emph{Invented continuations}: the model sometimes wrote the rest of the episode in one reply, with fake environment responses and a closing \act{finished} call; in one case, the agent submitted an answer at step~1 without ever seeing the data. We keep only the first Thought/Action pair. (ii)~\emph{Self-generated chat-template tokens}: the model occasionally appended tokens such as \texttt{<|eot\_id|>} or \texttt{<|end\_of\_turn|>}, probably imitating the \texttt{<|box\_start|>} syntax used in the prompt. These tokens were then typed into search boxes (Figure~\ref{fig:token}), and because each reply is fed back into the history, they recurred at every later step. Of the 305 task episodes we ran locally, 15 (4.9\%) were contaminated; 12 of the 13 scorable ones failed, and success was 7.7\% for contaminated episodes versus 41.7\% for clean ones. This comparison is observational; the offline replay below shows that the fix itself is safe. We strip such tokens, together with stray markup after the action (such as \texttt{<br>}), while preserving \texttt{<|box\_start|>}/\texttt{<|box\_end|>}. When we replayed all 7,705 recorded steps offline, the fix changed all 551 contaminated steps and none of the 7,154 clean ones.

\subsection{Execution: making actions land}
\label{sec:execution}

\paragraph{Coordinate-space alignment.} The reference implementation maps model coordinates to pixels with UI-TARS's \texttt{smart\_resize} convention, which assumes a 28-pixel patch grid~\cite{qin2025uitars,bai2025qwen25vl}. For a 1920$\times$1080 screenshot, this mapping is almost the identity, so the code treated the model's coordinates as full-resolution pixels. Yet across 1,312 clicks in our first full run, \emph{no} click fell to the right of $x{=}1440$ or below $y{=}810$, and on two separate tasks the ratio between the intended and the clicked position was 4/3. A calibration experiment on a synthetic target image with six known markers showed that the API was downscaling the image before the model saw it. When we sent a 1920$\times$1080 screenshot, the model's coordinates needed a correction factor of 1.324/1.350 in $x$/$y$, consistent with downscaling to about 1440$\times$810. When we sent 1440$\times$810, the factor was 1.010/0.993.

\begin{figure}[t]
\centering
\begin{minipage}[t]{0.49\columnwidth}\centering
\includegraphics[width=\linewidth]{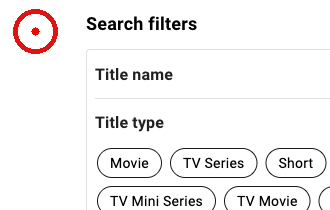}\\[-1pt]{\footnotesize (a) old mapping}
\end{minipage}\hfill
\begin{minipage}[t]{0.49\columnwidth}\centering
\includegraphics[width=\linewidth]{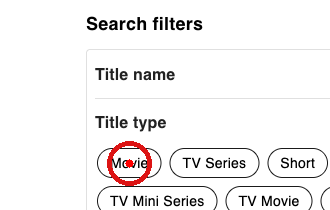}\\[-1pt]{\footnotesize (b) after alignment}
\end{minipage}
\caption{The same click before and after coordinate alignment (IMDb Advanced Title Search; red dot = executed click, ring added for visibility). (a) The model aimed at ``Movie'' at (287,400) in the downscaled image; the old mapping executed it almost unchanged at (285,396) on the 1920$\times$1080 page. (b) After alignment, the model's (284,396) is scaled by 4/3 to (379,528) and hits.}
\label{fig:coord}
\end{figure}

The fix is to downscale every screenshot ourselves to at most 1440$\times$810 and to map coordinates by the ratio of the \emph{sent} image to the \emph{actual} viewport, measured at runtime. On a real page, three targets that had been missed by 122--189\,px were hit within 1--5\,px (Figure~\ref{fig:coord}). The model had been accurate to within a few pixels all along; the error was entirely in the conversion. On a paired set of 24 dev tasks, success rose from 8/24 to 20/24 with the version containing this fix (12 fixed, 0 broken), and the share of tasks that hit the step limit fell from 67\% to 4\%. That version also included other changes; we attribute most of the gain to this fix because misplaced clicks are what had kept tasks looping until they hit the step limit. Because the baseline could still take the \act{goto} shortcut (\S\ref{sec:guardrails}), which inflated its score, the true gain is, if anything, larger.

\paragraph{Acting on elements, not just pixels.} Some actions cannot be delivered by a pixel click. A native \texttt{<select>} opens a popup that the browser draws outside the page, so it never appears in the screenshot; in one task, the model clicked the same dropdown nine times and gave up. \act{select} locates the element at the given point and sets the option directly, trying in order an exact match, a match after normalization (``Vietnam'' matches ``Viet Nam''), and a substring match. Text input has a similar problem. With \act{click}+\act{type}, keystrokes go to whichever element has focus, so a slightly missed click sends them to the wrong element or nowhere; in 15 of the dev70 tasks, the model re-entered the same input at least once. \act{fill} focuses the editable element at the given point, including inside iframes and under overlays, clears it, and types real keystrokes so that autocomplete lists still appear. For native date inputs, it converts the value to ISO format.

\paragraph{Keeping the prompt's promises.} The reference prompt told the model to end a \act{type} with \texttt{\textbackslash n} to submit, but its executor never pressed Enter; 79 of 356 \act{type} calls (22\%) had this form. The executor now presses Enter for them.

\begin{figure*}[t]
\centering
\begin{minipage}[b]{0.27\textwidth}\centering
\includegraphics[width=\linewidth]{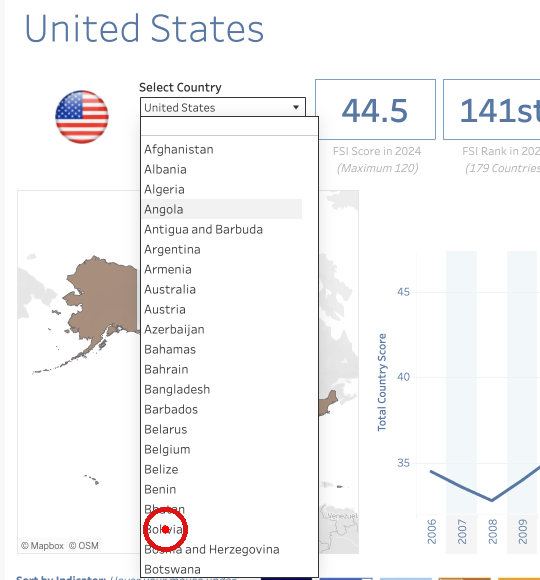}\\[-1pt]{\footnotesize (a) click inside the iframe}
\end{minipage}\hfill
\begin{minipage}[b]{0.70\textwidth}\centering
\includegraphics[width=\linewidth]{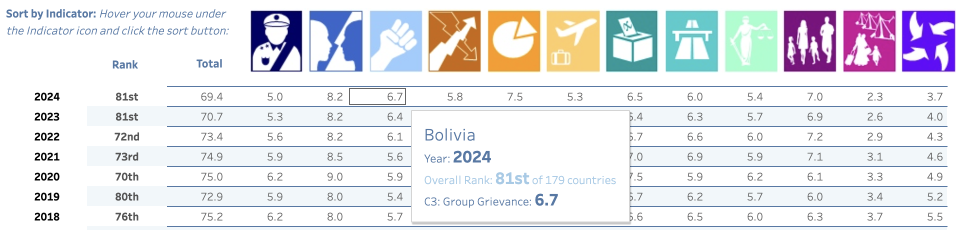}\\[-1pt]{\footnotesize (b) final screenshot at \act{finished(content='6.7')}}
\end{minipage}
\caption{Execution and feedback working together on a Tableau dashboard (Fragile States Index; task: Bolivia, 2024, Group Grievance). (a) The country list is inside an embedded frame at the bottom of the viewport. A first physical click on ``Bolivia'' had no effect. On the second, the child-frame fingerprint showed no change, so the executor dispatched a DOM click inside the frame, and the dashboard re-rendered for Bolivia. This task had never been solved in earlier runs with step limits of 40 or 80. (b) The answer is on screen, in the hover tooltip, when the agent finishes.}
\label{fig:iframe}
\end{figure*}

\paragraph{Iframe DOM fallback.} Embedded charts and document viewers, such as Flourish, Tableau, and SEC's inline XBRL viewer, swallow some physical clicks. A scripted probe that made no model calls showed that three physical clicks on a cross-origin iframe near the bottom of the viewport had no effect, while one DOM-level click worked. When an action lands in a child frame and neither frame changes (as detected by the feedback stage, \S\ref{sec:feedback}), the executor dispatches a complete sequence of DOM mouse events inside that frame (Figure~\ref{fig:iframe}). Frames from CAPTCHA providers are excluded: synthetic events there had no effect in our tests, and they are exactly what anti-bot systems look for.

\subsection{Feedback: reporting what actually happened}
\label{sec:feedback}
Before and after every action, the executor records a DOM fingerprint (URL, title, node count, text length, scroll offsets, focus) and compares the two. Unlike a pixel diff, the fingerprint is not fooled by ticking clocks or rotating carousels. If nothing changed, the model receives a feedback note naming the element under the click point and whether it is clickable. Two refinements were needed. First, the fingerprint originally covered only the main frame, so a \emph{successful} click inside an iframe was reported as ``no detectable change'', and the model abandoned an action that had worked. We now also fingerprint the child frame the action lands in, and we no longer count focus moving into an iframe as a change, since that alone would make every click on an iframe look successful. Second, the fingerprint records only the length of the page text, so a tooltip replaced by another of the same length went unnoticed; for hovers and for SVG/canvas targets, we also compare a hash of the page text. Actions also report their own results: \act{fill} reads back the box's actual value, and a failed \act{select} returns the list of available options, which the model has no other way to see.

Accurate feedback is necessary but not sufficient: on its own, it did not change the model's behavior (\S\ref{sec:negative}). We therefore pair it with budgets that the executor enforces (\S\ref{sec:guardrails}).

\subsection{Observation: showing the model what it needs}
\label{sec:observation}
The screenshot is the model's main view, so information that never reaches it cannot inform the answer. \act{hover} reveals values that exist only in tooltips on charts and maps. \act{find\_text} works like the browser's Ctrl+F: it scrolls the next match into view, searching child frames if the main frame has none, but it \emph{never} returns text. The answer must still be read from the screenshot, which is the condition under which the organizers approved the tool. \act{read\_text} returns up to 600 characters of DOM text near a label. The prompt restricts it to \emph{verifying} a value already read from the screenshot, not replacing it. Some sites hide digits behind custom fonts: the DOM holds rarely used code points that only the site's font renders as digits. \act{read\_text} replaces each such character with ``?'' and reports how many it replaced. Discarding every passage that contains such characters would not work: on one site, only 3.5\% of a line was obfuscated, and that 3.5\% was the value we needed. The screenshot size cap (\S\ref{sec:execution}) also helps here: an image sent at 1440$\times$810 is not resampled again by the API.

Tools help only if they are used at the right moment. We write each trigger as a rule in the prompt (``if your answer is a specific value, call \act{read\_text} once before \act{finished}'') rather than leaving it to the model's judgment. On six targeted tasks, rewriting the trigger for \act{find\_text} as a rule raised its use from 2 tasks to 3; \act{read\_text}, written as a rule from the start, was used in 5. Being used is not the same as helping, however: \act{read\_text} fixed none of those tasks (\S\ref{sec:negative}). The model is often confident even when it is wrong, so it cannot reliably decide for itself when to check.

\subsection{Guardrails: rules, budgets, and robustness}
\label{sec:guardrails}

\paragraph{Rules built into the harness.} An early version of the agent had a \act{goto(url)} action. An audit of its dev70 run found 692 \act{goto} calls: 18 of 70 tasks queried data APIs directly and 21 visited search engines, and 58\% of its correct answers came from such shortcuts. Rather than asking the model to behave, \agentname{} removes these paths from the action set, which is closed: there is no URL navigation, JavaScript, or HTTP access. \act{goto} and browser-level shortcuts (\texttt{ctrl+l}, \texttt{F5}, \ldots) are still parsed, but only so that the executor can reject them with an explanation. What the action set cannot rule out, the prompt covers with fourteen hard constraints, including: stay on the site, set named filters through the page's own controls, answer only from what was seen in this session (or submit an empty answer), and call \act{finished} only when the answer is visible on the \emph{current} screenshot with filters still applied. Started tasks are never retried; after a browser disconnect, the task ends as a recorded failure. In the last 24-task regression run before submission, every final page had been reached through the site's own interface, with no search-engine visits and no hand-built URLs. The organizers' manual review deducted nothing.

\paragraph{Budgets enforced by the executor.} Each task gets 80 steps and 1,800\,s. The two limits interact: each step includes a model call and page rendering, so the time limit is often reached first, and raising one limit without the other has little effect. Per-action budgets (Table~\ref{tab:actions}) are enforced by the executor, which rejects over-budget calls with an explanation. The reference implementation's scroll breaker ended a task after 10 consecutive scrolls. We changed it to end the task only if the page has also stopped changing, because the old rule had terminated a task that was legitimately reading a long table.

\paragraph{Robustness.} Model calls are retried on 403/429/5xx errors and timeouts, up to 6 attempts within 240\,s; previously, repeated 403s or connection errors ended 4 of 305 task episodes (1.3\%) at step~0. Tasks are dispatched from a shared queue rather than split statically, so a crashed worker loses one task rather than all of its remaining assigned tasks. After navigation, the executor waits between 1 and 8\,s, returning as soon as the page has settled, instead of a fixed 3\,s. Every exit path writes the required result files.

\section{Experiments}
\label{sec:exp}

\subsection{Setup}
\label{sec:expsetup}
With the organizers' explicit permission, we developed on the 100 publicly released Protocol~III tasks~\cite{dong2026webretrieverlargescalecomprehensivebenchmark}, split deterministically into dev70 and holdout30. We scored answers locally with exact, numeric (0.5\% relative tolerance), and date matching, plus an optional LLM judge for semantic equivalence. Sixteen of the 100 sites were unreachable from our network (\S\ref{sec:failure}), which leaves 25 scorable tasks in holdout30. On the same code version, these 25 tasks scored 52.0\% locally, and the hidden set scored 46.0 officially. The task sets differ, but the 6-point gap is within one binomial standard deviation (about 10 points for $n{=}25$), so we treat local numbers as a reasonable guide for relative comparisons.

\subsection{Official score trajectory}
Table~\ref{tab:official} lists our four official submissions on the hidden set. Each submission bundled several changes, so gains cannot be attributed to single components. The +10 at the second submission came mostly from finishing more tasks. The first evaluation run had finished only 79 of 100 tasks when it reached its time limit; at its own accuracy (31 of 79, 39.2\%), finishing 97 tasks, as the second run did, would have scored about 38. So roughly 7 of the 10 points came from completing more tasks and about 3 from answering more accurately.

\begin{table}[tb]
\centering\small
\caption{Official hidden-set scores in 2026 (answer accuracy). Manual review deducted no points from the final submission.}
\label{tab:official}
\begin{tabularx}{\columnwidth}{@{}lr>{\raggedright\arraybackslash}X@{}}
\toprule
Date & Score & Main additions \\
\midrule
Aug 29 & 31.0 & coordinate alignment, \act{select}, \act{hover}, \act{reload}; closed action set, hard constraints \\
Aug 31 & 41.0 & \act{find\_text}, \act{back}; shared-queue dispatch \\
Sep 1 & 46.0 & token stripping; \act{fill}, Enter on newline; \act{read\_text}; scroll-breaker fix \\
Sep 2 & \textbf{57.0} & iframe fallback; \act{find\_text}/\act{read\_text} search child frames; 80 steps; model-call retries \\
\bottomrule
\end{tabularx}
\end{table}

\subsection{Evidence for each stage}
A 100-task evaluation cost about \$170--230 and returned only an aggregate score, so we evaluated components on small targeted sets. Every paired comparison also included tasks that were \emph{already passing}, so that regressions would show up. Table~\ref{tab:ablation} groups the strongest evidence by stage. The sample sizes are small, and we report them to show direction, not statistical significance. Most rows repeat numbers from \S\ref{sec:method}; the rest are new. \act{hover}, introduced together with coordinate alignment, solved 4 of 5 tasks that need tooltip values. On 8 tasks, raising the step limit from 40 to 80 increased the number solved from 0--1 to 4, and 2 of those 4 needed more than 40 steps. The \act{reload} budget was used in 5 tasks: it recovered 2 that were stuck on loading pages, saved budget in 2 others, and hurt none.

\begin{table}[tb]
\centering\small
\caption{Component evidence on local dev tasks, grouped by stage. ``$+x/-y$'': tasks fixed / broken in a paired comparison. $n$ counts tasks unless noted.}
\label{tab:ablation}
\begin{tabularx}{\columnwidth}{@{}>{\raggedright\arraybackslash}X r >{\raggedright\arraybackslash}p{0.38\columnwidth}@{}}
\toprule
Component & $n$ & Before $\rightarrow$ after \\
\midrule
\multicolumn{3}{@{}l}{\textit{Parsing}} \\
Token stripping (offline replay; $n$ = steps) & 7{,}705 & 551/551 fixed; 0 clean steps changed \\
\multicolumn{3}{@{}l}{\textit{Execution}} \\
Coordinate alignment$^\ast$ & 24 & 8 $\rightarrow$ 20 (+12/$-$0) \\
\quad on a real page ($n$ = targets) & 3 & miss by 122--189\,px $\rightarrow$ 1--5\,px \\
\act{fill} + Enter on newline & 10 & 2 $\rightarrow$ 4 (+2/$-$0); 12\% fewer steps \\
\multicolumn{3}{@{}l}{\textit{Execution and feedback}} \\
Iframe fallback (Tableau) & 1 & never solved $\rightarrow$ pass in 25 steps \\
\multicolumn{3}{@{}l}{\textit{Observation}} \\
\act{hover} (with alignment) & 5 & 0 $\rightarrow$ 4 \\
\act{find\_text} (single task) & 1 & fail at 40 steps $\rightarrow$ pass in 12 \\
\multicolumn{3}{@{}l}{\textit{Guardrails}} \\
Hotkey block-list$^\dagger$ & 10 & 0 $\rightarrow$ 2 (+2/$-$0) \\
Step limit 40 $\rightarrow$ 80 & 8 & 0--1 $\rightarrow$ 4 (2 needed $>$40 steps) \\
Scroll-breaker criterion & 3 & 2 tasks recovered \\
\act{reload} budget & 5 & 2 recovered, 2 saved budget, 0 hurt \\
\bottomrule
\end{tabularx}
\par\vspace{3pt}
{\raggedright\footnotesize $^\ast$Version-level comparison; alignment was the main change, and the baseline still had the \act{goto} shortcut. $^\dagger$Shipped together with the feedback notes and a later-removed \act{goto}.\par}
\end{table}

\subsection{Negative results}
\label{sec:negative}
Several changes that did not work shaped the final design:
\begin{itemize}
\item \textbf{Feedback alone does not change behavior.} Adding ``no detectable change'' notes left 10 idle-looping tasks at 0/10, with the same 400 total steps. The notes reached the model, and its thoughts acknowledged them, yet it kept repeating the action. This is why the executor enforces budgets.
\item \textbf{A heuristic idle breaker} (stop after 6 no-change steps) terminated a task that would otherwise have succeeded and recovered none, so we reverted it.
\item \textbf{\act{read\_text}} was used in 5 of 6 targeted tasks but fixed none of them. The model looked up short, ambiguous labels (one label matched 124 elements). It also broke none, so we kept it, capped at five calls per task.
\item \textbf{An early step-limit test} (40$\rightarrow$65) showed no gain. Repeated on current code (40$\rightarrow$80), the comparison showed a real gain (Table~\ref{tab:ablation}): the first test had been run on an outdated version, a reminder to repeat ablations after upstream fixes.
\item \textbf{The metric we chose in advance} for \act{fill} (fewer repeated inputs) did not move (6$\rightarrow$6). The improvement came through a different route: 12\% fewer steps.
\end{itemize}

\subsection{Failure analysis}
\label{sec:failure}
Table~\ref{tab:buckets} groups all 100 local tasks by their best outcome across all our runs, so the count of correct answers is an upper bound for any single version. Many of the remaining errors are not the agent's. In 13 of these public tasks (not the hidden evaluation set), the reference answer no longer matches the live site because the data has been updated, or it uses a different metric definition (for example, 100-year versus 20-year global warming potential). Sixteen sites were unreachable from our network. The agent's own errors are either capability gaps, meaning unsolved execution and observation problems such as canvas-only widgets and dense charts, or decision errors, which a verifier and task-type knowledge might address (\S\ref{sec:future}). For the maintenance of WebRetriever~\cite{dong2026webretrieverlargescalecomprehensivebenchmark} and similar live-site benchmarks, we suggest recording each reference answer together with a timestamped evidence screenshot.

\begin{table}[t]
\centering\small
\caption{Best outcome per local task (100 public Protocol~III tasks; approximate; some tasks fall into more than one bucket).}
\label{tab:buckets}
\begin{tabularx}{\columnwidth}{@{}Xr@{}}
\toprule
Bucket & Tasks \\
\midrule
Answered correctly (best across runs) & $\sim$55 \\
Reference answer outdated or defined differently & 13 \\
Site unreachable from our network & 16 \\
Structural: image-grid CAPTCHA, canvas-only UI, answer needs a typed URL or spans several pages & 6 \\
Capability gaps (iframe, chart, control) & $\sim$4 \\
Path not found / wrong decision & 3 \\
\bottomrule
\end{tabularx}
\end{table}

\subsection{Cost}
The average step consumed about 10.6k input and 127 output tokens; input exceeded output 83-fold, almost entirely because the five most recent screenshots are re-sent at every step. A 100-task run cost about \$170--230. Prompt caching was unavailable through the mandated OpenAI-compatible endpoint.

\section{Discussion}
\label{sec:discussion}

\subsection{Design principles}
We expect five principles, each supported by the evidence above, to apply beyond this challenge:
\begin{enumerate}
\item \textbf{Find the failing stage before blaming the model.} Several failures that looked like reasoning errors, such as clicking empty space (\S\ref{sec:execution}) or abandoning an iframe click that had worked (\S\ref{sec:feedback}), were execution or feedback failures.
\item \textbf{Define each action consistently everywhere it appears:} in the prompt that describes it, the parser, the executor, and the feedback it produces. If any of these is missing, the action fails silently: \act{hover} could be executed but was never described to the model; the Enter on \texttt{\textbackslash n} was promised but never pressed; the parser did not strip \texttt{<|eot\_id|>}. Automated tests now check this consistency.
\item \textbf{Enforce budgets in the executor, not the prompt.} Prompt instructions alone did not stop repeated actions; hard limits with an explanation did.
\item \textbf{Write trigger conditions as rules}, not as invitations to use a tool ``when needed''.
\item \textbf{Add a compound action only when it removes a class of silent failure.} \act{fill} was added because keyboard focus cannot be verified from a screenshot, not to save steps. That its measured gain came through fewer steps anyway (\S\ref{sec:negative}) shows why the mechanism and the outcome should be measured separately.
\end{enumerate}

\subsection{Limitations}
\label{sec:limitations}
Our evidence has clear limits. (i)~\emph{One model.} All results use \texttt{claude-opus-4-6}. The four-stage view is model-agnostic, but we have not shown that each fix transfers, and some (coordinate alignment, token stripping) address problems specific to this model and its API. (ii)~\emph{No same-model baseline on the hidden set.} We never submitted the reference implementation itself, so the 31.0 starting point already includes several of our fixes; locally, the reference implementation with our model solved none of a 10-task subset. (iii)~\emph{Small samples.} Component evidence comes from targeted sets of 1--24 tasks and shows direction, not significance. (iv)~\emph{Live websites.} Sites change and some were unreachable from our network, so repeated runs of the same version vary, and each official score comes from a single run. (v)~\emph{Development data.} We developed on the public Protocol~III tasks, so our local numbers are not comparable to leaderboard scores~\cite{webretriever_leaderboard}.

\subsection{Future work}
\label{sec:future}
We list future directions roughly in order of expected payoff relative to effort:
\begin{itemize}
\item \textbf{Routing steps to different models.} Steps differ in what they need: locating a control, reading a dense table, deciding when to stop. Each step can go to a large or small multimodal model, a rule, or a classical vision routine, while the four stages and the guardrails stay the same.
\item \textbf{Classical computer vision for feedback and observation.} OCR to cross-check values read from screenshots; perceptual-hash change detection that catches canvas and image updates invisible to the DOM fingerprint; template matching for recurring widgets. The aim is an agent that keeps looking until the evidence is clear.
\item \textbf{An independent verifier.} Before \act{finished}, a separate model call with a fresh context, seeing only the task, the final screenshot, and the candidate answer, checks the answer, targeting wrong-row and wrong-table errors.
\item \textbf{Native API with prompt caching and tool-use actions.} Outside the challenge's API constraints, this would cut cost by an estimated 55--70\% and remove free-text parsing, and with it most parsing failures.
\item \textbf{Hybrid observation.} A list of interactive elements with coordinates, shown alongside the screenshot. This should be A/B-tested, since SeeAct~\cite{zheng2024seeact} shows that hybrids are not automatically better.
\item \textbf{Download-and-parse} for PDF and spreadsheet sources, and \textbf{task-type skills}: short per-category checklists plus per-site operating notes, similar to the operational documentation of WebRetriever's Protocol~II~\cite{dong2026webretrieverlargescalecomprehensivebenchmark}.
\item \textbf{Synthetic tasks and fine-tuning.} Tasks with programmatic ground truth can be generated on the same sites, and teacher trajectories distilled into a smaller open model, to cut cost and internalize the prompt rules.
\end{itemize}

\section{Conclusion}
In our experience, when a vision-based web agent fails on a live website, the cause often lies not in its reasoning but in how its decisions are carried out and reported back. \agentname{} treats every step as a round trip through four stages (parsing, execution, feedback, and observation) and surrounds the loop with guardrails. Making these stages reliable raised its WebRetriever Challenge score from 31.0 to 57.0 across four submissions and placed it 2nd. Because the four-stage view does not depend on the model, it should carry over to stronger models, to combinations of models, and to new environments. Our main advice to builders of web agents: before blaming the model, check that its decisions actually reach the page, and that the page's response actually reaches the model.

\section*{Code Availability}
The source code of \agentname{} is publicly available on GitHub:\\
\url{https://github.com/jianganghan/WebFovea}

\section*{Acknowledgments}
We thank the organizers of the WebRetriever Challenge 2026 at Mininglamp Technology and the authors of WebRetriever~\cite{dong2026webretrieverlargescalecomprehensivebenchmark} for the benchmark, the evaluation infrastructure, and timely answers to our questions about the rules.

\section*{Disclosure of AI Use}
Separately from the agent's own model (\texttt{claude-opus-4-6}, \S\ref{sec:setting}), the author used Claude Code (Anthropic), an AI coding assistant, throughout this project: to implement and test agent components, to analyze experiment logs, and to help draft this report. The author conceived the approach and set the research direction; decided which failures to investigate and which changes to adopt, keep, or revert; designed the experiments and judged what their results did and did not show; checked the agent's behavior against its trajectories; and reviewed all code, analyses, and text. The author takes full responsibility for the content.

\bibliographystyle{unsrtnat}
\bibliography{refs}

@misc{dong2026webretrieverlargescalecomprehensivebenchmark,
    title={{WebRetriever}: A Large-Scale Comprehensive Benchmark for Efficient Web Agent Evaluation},
    author={Wei Dong and Tianyu Fu and Zhe Yu and Hanning Wang and Anyang Su and Zhizhou Fang and Yuyang Chen and Shuo Wang and Minghui Wu and Ping Jiang and Zhen Lei and Chenxu Zhao},
    year={2026},
    eprint={2607.06118},
    archivePrefix={arXiv},
    primaryClass={cs.CV},
    howpublished={arXiv:2607.06118},
}

@misc{webretriever_leaderboard,
    title={{WebRetriever} Leaderboard},
    author={{WebRetriever Team}},
    year={2026},
    howpublished={\url{https://mininglamp-ai.github.io/WebRetriever/}},
    note={Accessed September 2026},
}

@misc{webretriever_challenge,
    title={{WebRetriever Challenge 2026}},
    author={{Mininglamp Technology}},
    year={2026},
    howpublished={\url{https://mininglamp-ai.github.io/WebRetriever_Challenge/}},
}

@inproceedings{zhou2024webarena,
    title={{WebArena}: A Realistic Web Environment for Building Autonomous Agents},
    author={Zhou, Shuyan and Xu, Frank F. and Zhu, Hao and Zhou, Xuhui and Lo, Robert and Sridhar, Abishek and Cheng, Xianyi and Ou, Tianyue and Bisk, Yonatan and Fried, Daniel and Alon, Uri and Neubig, Graham},
    booktitle={International Conference on Learning Representations (ICLR)},
    year={2024},
}

@inproceedings{deng2023mind2web,
    title={{Mind2Web}: Towards a Generalist Agent for the Web},
    author={Deng, Xiang and Gu, Yu and Zheng, Boyuan and Chen, Shijie and Stevens, Samuel and Wang, Boshi and Sun, Huan and Su, Yu},
    booktitle={Advances in Neural Information Processing Systems (NeurIPS)},
    year={2023},
}

@inproceedings{he2024webvoyager,
    title={{WebVoyager}: Building an End-to-End Web Agent with Large Multimodal Models},
    author={He, Hongliang and Yao, Wenlin and Ma, Kaixin and Yu, Wenhao and Dai, Yong and Zhang, Hongming and Lan, Zhenzhong and Yu, Dong},
    booktitle={Proceedings of the Annual Meeting of the Association for Computational Linguistics (ACL)},
    year={2024},
}

@inproceedings{zheng2024seeact,
    title={{GPT-4V(ision)} is a Generalist Web Agent, if Grounded},
    author={Zheng, Boyuan and Gou, Boyu and Kil, Jihyung and Sun, Huan and Su, Yu},
    booktitle={Proceedings of the 41st International Conference on Machine Learning (ICML)},
    series={Proceedings of Machine Learning Research},
    volume={235},
    pages={61349--61385},
    year={2024},
}

@misc{abuelsaad2024agente,
    title={{Agent-E}: From Autonomous Web Navigation to Foundational Design Principles in Agentic Systems},
    author={Abuelsaad, Tamer and Akkil, Deepak and Dey, Prasenjit and Jagmohan, Ashish and Vempaty, Aditya and Kokku, Ravi},
    year={2024},
    eprint={2407.13032},
    archivePrefix={arXiv},
    howpublished={arXiv:2407.13032},
}

@misc{browseruse2024,
    title={{Browser Use}: Enable {AI} to Control Your Browser},
    author={M{\"u}ller, Magnus and {\v{Z}}uni{\v{c}}, Gregor},
    year={2024},
    publisher={GitHub},
    howpublished={\url{https://github.com/browser-use/browser-use}},
}

@misc{qin2025uitars,
    title={{UI-TARS}: Pioneering Automated {GUI} Interaction with Native Agents},
    author={Qin, Yujia and Ye, Yining and Fang, Junjie and Wang, Haoming and Liang, Shihao and others},
    year={2025},
    eprint={2501.12326},
    archivePrefix={arXiv},
    howpublished={arXiv:2501.12326},
}

@misc{gemini25,
    title={Gemini 2.5: Pushing the Frontier with Advanced Reasoning, Multimodality, Long Context, and Next Generation Agentic Capabilities},
    author={Comanici, Gheorghe and Bieber, Eric and Schaekermann, Mike and Pasupat, Ice and Sachdeva, Noveen and others},
    year={2025},
    eprint={2507.06261},
    archivePrefix={arXiv},
    howpublished={arXiv:2507.06261},
}

@misc{anthropic2024computeruse,
    title={Introducing Computer Use, a New {Claude 3.5 Sonnet}, and {Claude 3.5 Haiku}},
    author={{Anthropic}},
    year={2024},
    howpublished={\url{https://www.anthropic.com/news/3-5-models-and-computer-use}},
}

@misc{bai2025qwen25vl,
    title={{Qwen2.5-VL} Technical Report},
    author={Bai, Shuai and Chen, Keqin and Liu, Xuejing and Wang, Jialin and Ge, Wenbin and others},
    year={2025},
    eprint={2502.13923},
    archivePrefix={arXiv},
    howpublished={arXiv:2502.13923},
}

@misc{yang2024sweagent,
    title={{SWE-agent}: Agent-Computer Interfaces Enable Automated Software Engineering},
    author={Yang, John and Jimenez, Carlos E. and Wettig, Alexander and Lieret, Kilian and Yao, Shunyu and Narasimhan, Karthik and Press, Ofir},
    year={2024},
    eprint={2405.15793},
    archivePrefix={arXiv},
    howpublished={arXiv:2405.15793},
}

@misc{cheng2024seeclick,
    title={{SeeClick}: Harnessing {GUI} Grounding for Advanced Visual {GUI} Agents},
    author={Cheng, Kanzhi and Sun, Qiushi and Chu, Yougang and Xu, Fangzhi and Li, Yantao and Zhang, Jianbing and Wu, Zhiyong},
    year={2024},
    eprint={2401.10935},
    archivePrefix={arXiv},
    howpublished={arXiv:2401.10935},
}

@misc{yang2023setofmark,
    title={{Set-of-Mark} Prompting Unleashes Extraordinary Visual Grounding in {GPT-4V}},
    author={Yang, Jianwei and Zhang, Hao and Li, Feng and Zou, Xueyan and Li, Chunyuan and Gao, Jianfeng},
    year={2023},
    eprint={2310.11441},
    archivePrefix={arXiv},
    howpublished={arXiv:2310.11441},
}

@misc{xie2024osworld,
    title={{OSWorld}: Benchmarking Multimodal Agents for Open-Ended Tasks in Real Computer Environments},
    author={Xie, Tianbao and Zhang, Danyang and Chen, Jixuan and Li, Xiaochuan and Zhao, Siheng and Cao, Ruisheng and Hua, Toh Jing and Cheng, Zhoujun and Shin, Dongchan and Lei, Fangyu and Liu, Yitao and Xu, Yiheng and Zhou, Shuyan and Savarese, Silvio and Xiong, Caiming and Zhong, Victor and Yu, Tao},
    year={2024},
    eprint={2404.07972},
    archivePrefix={arXiv},
    howpublished={arXiv:2404.07972},
}

\clearpage
\onecolumn
\appendix
\section{Component Inventory}
\label{app:inventory}
Table~\ref{tab:inventory} lists every component of \agentname{}. For each one, it gives the stage it belongs to (\S\ref{sec:links}) and where it came from: \emph{inherited} unchanged from the reference implementation, \emph{modified} from it, or \emph{new}. The table also includes components that the main text mentions only briefly, such as \act{back}, \act{reload}, \act{click\_many}, and the task dispatcher.

\begin{table}[h]
\centering\small
\caption{Component inventory. Not listed (inherited unchanged): CDP connection and authentication, click/scroll/drag primitives, randomized keystroke delay, no-cache request headers.}
\label{tab:inventory}
\begin{tabularx}{\textwidth}{@{}l>{\raggedright\arraybackslash}X l @{\hspace{2.5em}} l>{\raggedright\arraybackslash}X l@{}}
\toprule
Stage & Component & Origin & Stage & Component & Origin \\
\midrule
Parsing & first-action truncation & new & Observation & 5-screenshot window + full text history & inherited \\
Parsing & special-token / trailing-markup stripping & new & Observation & screenshot size cap (1440$\times$810) & new \\
Parsing & answer extraction from \act{finished} & modified & Observation & \act{hover} exposed to the model & modified \\
Execution & coordinate-space alignment & new & Observation & \act{find\_text} (frame-aware, returns no text) & new \\
Execution & \act{select} for native dropdowns & new & Observation & \act{read\_text} with de-obfuscation & new \\
Execution & \act{fill} with date handling & new & Observation & rule-based tool triggers & new \\
Execution & Enter on trailing \texttt{\textbackslash n} & new & Guardrails & closed action set; \act{goto} rejected & new \\
Execution & iframe DOM fallback & new & Guardrails & browser-level hotkey block-list & new \\
Execution & \act{click\_many} (image-grid CAPTCHA only) & new & Guardrails & 14 hard constraints & new \\
Execution & \act{back}, \act{reload} & new & Guardrails & executor budgets; 80 steps / 1,800\,s & modified \\
Feedback & DOM fingerprint + notes & new & Guardrails & scroll breaker (requires no change) & modified \\
Feedback & child-frame fingerprint & new & Guardrails & no task retries; model-call retries & modified \\
Feedback & tooltip text hash & new & Guardrails & shared-queue dispatch; adaptive settle wait & modified \\
Feedback & \act{fill} read-back; \act{select} option list & new & & & \\
\bottomrule
\end{tabularx}
\end{table}

\end{document}